\documentclass[conference]{IEEEtran}
\IEEEoverridecommandlockouts
\usepackage{cite}
\usepackage[numbers,sort&compress]{natbib}
\usepackage{amsmath,amssymb,amsfonts}
\usepackage{algorithmic}
\usepackage{graphicx}
\usepackage{textcomp}
\usepackage{hyperref}
\usepackage[table,xcdraw]{xcolor}
\def\BibTeX{{\rm B\kern-.05em{\sc i\kern-.025em b}\kern-.08em
    T\kern-.1667em\lower.7ex\hbox{E}\kern-.125emX}}
\begin{document}

\title{Multi-modal Speech Transformer Decoders: When Do Multiple Modalities Improve Accuracy?}

\author{
\IEEEauthorblockN{
 Yiwen Guan, 
 Viet Anh Trinh, 
 Vivek Voleti, 
 Jacob Whitehill
}
\IEEEauthorblockA{
Worcester Polytechnic Institute \\
\{yguan2, vtrinh, vsvoleti, jrwhitehill\}@wpi.edu
}
}

\maketitle


\begin{abstract}
Decoder-only discrete-token language models have recently achieved significant success in automatic speech recognition. However, systematic analyses of how different modalities impact performance in specific scenarios remain limited. In this paper, we  investigate the effects of multiple modalities on speech recognition accuracy on both synthetic and real-world datasets. Our experiments suggest that: (1) Integrating more modalities can increase accuracy but the benefit depends on the amount of auditory noise. We also show for the first time the benefit of combining audio, image context, and lip information in one speech recognition model. (2) Images as a supplementary modality for speech recognition provide their greatest benefit at moderate audio noise levels; moreover, they exhibit a different trend compared to inherently synchronized modalities like lip movements. (3) Performance improves on both synthetic and real-world datasets when the most relevant visual information is filtered as a preprocessing step.
\end{abstract}



\section{Introduction}
\label{sec:intro}
When developing multi-modal automatic speech recognition (ASR) systems, when do the different input modalities (speech, visual context, lip movements of the speaker, etc.) help, and under what conditions is the benefit of each modality the strongest? 
The past few years have seen rapidly growing interest in ASR based on decoder-only discrete-token language models (e.g., \cite{chang2024exploring,team2023gemini,trinh2024discrete}). Such models  are attractive in part due to their ability to accept multi-modal inputs (e.g., audio, text, images) and generate multi-modal outputs (e.g., text tokens for ASR, audio tokens for speech-to-speech translation, etc.). 
The ability to process multiple input streams, such as audio that was spoken, lip movements of the speaker, an image of what was spoken about, etc.,
can enhance robustness of multi-modal ASR systems under challenging conditions, such as noisy environments. 
Advantages of such models include the ability to pretrain on large-scale audio or audio-visual datasets \cite{radford2023robust, afouras2018deep, shillingford2018large}, and to leverage advanced language understanding capabilities from LLMs \cite{achiam2023gpt} to perform multi-modal tasks. Despite recent development of multi-modal large language models (MLLMs) for speech processing (e.g., Google Gemini),  studies that systematically investigate how different modalities impact ASR performance are scarce. 

In this work, we systematically analyze the impact of different input modalities (speech audio, visual context, lip movements, text from optical character recognition) in multimodal ASR models across a range of noise conditions. Our goal is to identify noise regimes under which the modalities are complementary to each other. As a motivating example, if a speaker is presenting a Powerpoint slide but the speech signal is weak (due to unclear pronunciation or background noise), then the visual information of the slide could be used to help transcribe what the speaker said -- \emph{as long as} the speech was at least clear enough that a correspondence between the visual and auditory information can be established. Moreover, if the speaker's lips are visible, then lip-reading could potentially boost accuracy as well. On the other hand, the use of additional modalities might actually hurt performance due to a longer input length. Moreover, the fact that some modalities (e.g., speech \& lips) are \emph{synchronized} with each other whereas others (e.g., speech \& image context) are not, may make it more difficult for the model to establish a correspondence between information sources and require different processing architectures.

Our paper investigates the following research questions:
(1) Do additional modalities always help ASR accuracy?
 (2) Does each modality provide a uniform accuracy boost across all noise levels? 
 (3) How does irrelevant visual information affect ASR performance? 
To this end, we introduce a synthesized, multi-modal dataset (3-Equations) that is highly controllable, on which we can conveniently simulate various tailored situations.
This dataset focuses on mathematical scenarios, as specific educational applications are part of the motivation for this work.
We further experiment on the real-world SlideAVSR dataset \cite{wang2024slideavsr}.
In addition to these experiments, we also 
extend the MLLM in \cite{trinh2024discrete} to support more input modalities and demonstrate for the first time how audio, image context, and lip information can be combined in one model to increase ASR performance. Beyond multi-modal ASR, our findings have potential applications in areas like video conferencing, and educational AI partners \cite{https://doi.org/10.1111/bjet.13518}, where multi-modal inputs improve noisy environment performance and aid understanding in technical contexts.


\begin{figure*}[ht]
    \centering
    \includegraphics[width=\linewidth]{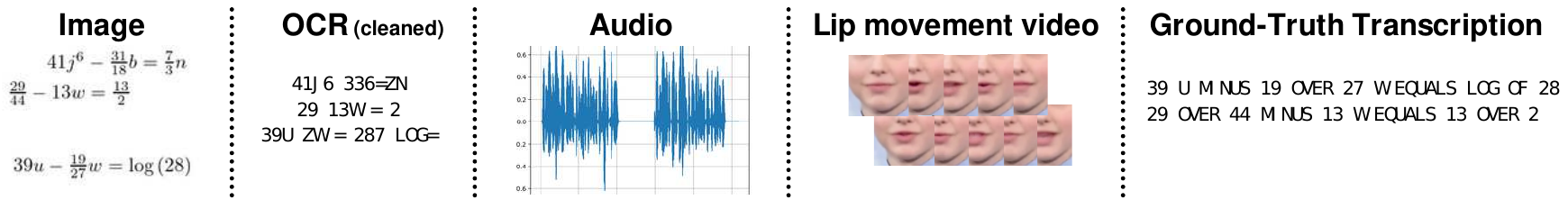}
    \vspace{-1.8em}
    \caption{An example of the 3-Equations dataset. From left to right: the image sample shows 3 mathematical equations; the OCR texts are extracted with EasyOCR and cleaned to retain only numbers, letters, and operators; the audio contains randomly reading 2 out of the 3 equations aloud; the lip movement video displays a lip region reading the corresponding equation sentences; the ground-truth transcription is the plain-text translation (label) of the speech. In this example, the speech reads the third and the second equations in order.}
    \vspace{-1em}
    \label{fig:eq_dataset}
\end{figure*}

\section{Related Work}
\label{sec:res}
\textbf{Audio-visual speech recognition}: \ 
Despite the tremendous success of ASR systems over the past decade, quality and reliability issues still exist in acoustically noisy environments \cite{math11122665}. One strategy to mitigate such noise is to harness complementary visual information, e.g., of the speaker's lips  \cite{ryumin2023audio,ma2021end,song2022multimodal,burchi2023audio,afouras2018deep}. 
A state-of-the-art approach is AV-HuBERT \cite{shilearning}: it learns audio-visual speech representations by feature clustering and masked prediction. By extending noise augmentation to AV-HuBERT pre-training, downstream models attain better robustness to acoustic noise \cite{shi2022robust}. Other works \cite{ma2023auto, hong2023watch} explore the performance gap between audio-only and audio-visual models in noisy conditions and show that the gap becomes larger as the noise levels increase. Instead of harnessing visual representations of the lips, some works such as SlideSpeech \cite{wang2024slidespeech} and SlideAVSR \cite{wang2024slideavsr}  have explored using optical character recognition (OCR) to extract information from the subject matter (lecture slides). To leverage the extracted text, SlideSpeech uses cross-attention to combine speech embeddings and contextual phrase embeddings for contextual ASR; SlideAVSR proposes DocWhisper which provides the texts to Whisper as prompts. Both works demonstrate performance improvement by integrating OCR texts.

\textbf{Multi-modal large language models}: \ Recognizing the powerful language generation, zero-shot transfer, and contextual learning capabilities of large language models (LLMs), significant efforts have been made to harness the knowledge from LLM pre-training to empower multi-modal tasks \cite{zhang2024mm}. Research works such as Qwen-Audio \cite{chu2023qwen} and LLaVA \cite{liu2024visual} have focused on multi-modal comprehension. Meanwhile, the outstanding generative capability has also inspired many works to extend unimodal LLMs to perform multi-modal generation, such as MiniGPT-5 \cite{zheng2023minigpt} and NExT-GPT \cite{wunext}, leading to the emergence of multi-modal large language models (MLLMs). Even so, only a few works incorporate visual speech modeling with LLMs, leaving this a relatively unexplored area. VSP-LLM \cite{yeo2024visual} integrates LLMs into visual speech processing, enhancing context modeling for tasks like visual speech recognition (VSR) and translation (VST). Llama-AVSR \cite{cappellazzo2024large} enables an LLM to perform ASR, VSR, and AVSR tasks by leveraging pre-trained audio-visual encoders. However, both works primarily concentrate on tasks like visual speech processing or AVSR while not exploring combinations with other modalities.


\section{Datasets}
\label{sec:dataset}
\vspace{-0.3em}

\subsection{3-Equations}
To systematically explore the interaction between modalities, we synthesized a multi-modal dataset consisting of images, audio clips, and lip movements. Each example contains multiple randomly generated mathematical equations -- reminiscent of the kinds of visual content that appear in lecture slides. The motivation is to design a dataset that forces the model to exploit information across modalities while allowing us to introduce complexity in a controlled manner. In this way, we can study how the model relies on information from each modality separately, especially when some modalities are insufficient or corrupted. An example from this dataset is illustrated in Fig.~\ref{fig:eq_dataset}. To simulate an insufficient auditory modality scenario, each audio sample only reads out two of the three equations randomly from the image. This setting encourages the model to rely on both visual and auditory modalities: without auditory information, the model cannot find the correct spoken equations; without visual information, the model will fail in a noisy environment.

The dataset consists of 10,000 examples, each containing one image sample with its OCR text, one audio sample, and one lip movement video sample. Specifically, the images depict three mathematical equations, involving operations such as addition, subtraction, logarithms, fractions and exponentiation, with each image sized $450 \times 200$. The audio part contains 25.2 hours of synthesized speech, averaging 9 seconds per sample, with 20.2 hours for training and 2.5 hours each for development and testing. The lip movement videos are generated from the synthesized audio at 25 FPS using a static portrait image. To create the dataset, we employ pyttsx3 \cite{pyttsx3github} to generate speech utterances, latex command to produce equation images, EasyOCR \cite{easyocr} to obtain OCR texts, and Wav2Lip \cite{prajwal2020lip} to generate lip-synced videos.

One advantage of creating this dataset is that, we can freely simulate anticipated situations, such as when multi-modal fusion is important. Hence, we added noise from the MUSAN dataset \cite{snyder2015musan} to the second half of each equation utterance at varying signal-to-noise ratios (SNR). This process produces a dataset we refer to as ``2-noise''. The first half of each utterance remains clean, allowing the model to leverage this ``incomplete'' clean auditory information to locate the correct equation in the image, and subsequently complete the speech transcription with clean visual information.

\begin{figure}[t]
    \centering    \includegraphics[width=\columnwidth]{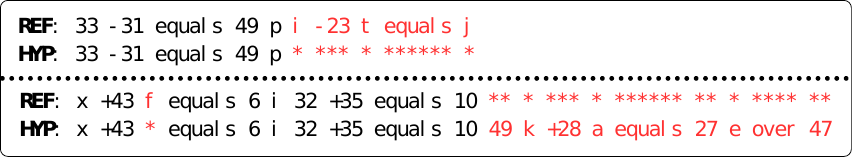}
    \vspace{-1.5em}
    \caption{Examples of Gemini's output on the 3-Equations 2-noise. For a fair comparison, we ``help'' Gemini spell out some words it recognizes as symbols. REF stands for reference, HYP stands for model's hypothesis.}
    \label{fig:gemini_example}
    \vspace{-1.3em}
\end{figure}

Another advantage of using this dataset is that, given that our image data consists of mathematical symbols, the visual information could be difficult for standard image codecs to extract. This allows us to study the impact of an imperfect image codec on the ASR accuracy. In particular, if we consider an ``oracle OCR'' that perfectly transcribes the 3 equation sentences in the image (but does not select which 2 out of the 3 were actually spoken), then the sequence: no vision, image encoding, real OCR, oracle OCR can be considered as an ascending order of image representation quality. Therefore, we can systematically explore how the quality of the visual modality, which is a supplementary modality for the task, affects speech recognition performance.

Interestingly, we identify a particular failure mode of Gemini when feeding 3-Equations 2-noise data into Gemini-1.5 Flash for transcription. Examples of Gemini's output are shown in Fig.~\ref{fig:gemini_example}. It appears to overly rely on a single modality, whereby prone to generate transcriptions for 1 or 3 equations instead of the correct 2 spoken equations. This finding also illustrates how this synthetic dataset can be useful for analyzing even very powerful models.

\subsection{SlideAVSR}
To generalize our findings to the real world, we leverage SlideAVSR \cite{wang2024slideavsr}, an audio-visual dataset of paper explanation videos collected from YouTube. The dataset provides manually transcribed speech, synchronized visual slides, and preprocessed OCR keywords. Since the videos contain many AI technical terms, accurate transcription is difficult without referring to the slides, which makes the dataset useful for multi-modal experiments. The dataset comprises 245 hours of audio data, with 195 hours allocated for training, 20 hours for development, and 30 hours for testing.
Since most of the videos in SlideAVSR are recorded from the presenter's laptop, the audio quality is generally very high. To simulate a broader range of noise conditions, we thus also add noise to the clean audio at different SNRs.


\section{Experimental Setup}
\label{sec:exp}

\begin{figure}[t]
\centering
    \includegraphics[width=\columnwidth]{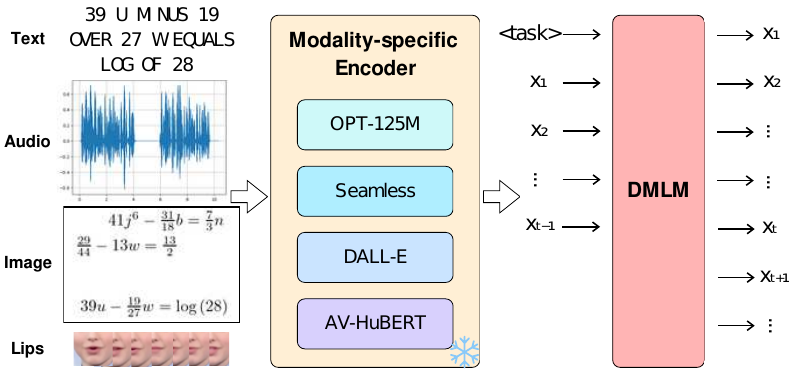}
    \vspace{-1.5em}
    \caption{An overview of Discrete Multimodal Language Model (DMLM). The inputs are encoded by modality-specific encoders, and the encoded multi-modal tokens are concatenated to a task prompt, and then passed to DMLM for processing in next-token prediction fashion.}
    \vspace{-1em}
    \label{fig:dmlm}
\end{figure}

\subsection{Model}
Our multi-modal speech recognition experiments are based on a Discrete Multi-modal Language Model (DMLM, depicted in Fig.~\ref{fig:dmlm}) \cite{trinh2024discrete}, which is a discrete token-based Transformer decoder model using OPT \cite{zhang2022opt} as the backbone. DMLM tokenizes the input data by frozen modality-specific encoders to form a discrete token sequence. In particular, DMLM uses the Seamless codec \cite{barrault2023seamlessm4t} to convert audio waveforms to discrete speech tokens, and the DALL-E encoder \cite{ramesh2021zero} to convert images to discrete image tokens. In addition, to extend DMLM to accept lip movement input, we employ AV-HuBERT to obtain lip tokens as well as text tokens representing the lip-reading hypothesis \cite{shilearning,shi2022robust}. The input token sequence is concatenated to a task description (e.g. ASR, lip-to-text) and then passed to DMLM for processing. 
Thanks to the attention mechanism, DMLM is able to learn how to combine information from different modality tokens. For example, in Fig.~\ref{fig:dmlm}, when the speaker is saying ``thirty-nine", the model learns to attend to the corresponding audio input (i.e. the beginning of the first audio segment), image input (i.e. the bottom left of the image), and lip input (i.e. the beginning of the lip movements) to complete the transcription.
Due to the parallelism of the input tokens from all modalities, the multi-modality can have a mild impact on the time complexity.

\subsection{Model Configurations and Training Details}
Following prior work \cite{trinh2024discrete}, we employ the OPT model with 125M parameters as the backbone LLM of DMLM. To enable DMLM to solve multi-modal problems, we fine-tune DMLM on a mixture of multi-modal tasks, including speech recognition, speech translation, image generation, and image captioning, using LibriSpeech-960 \cite{panayotov2015librispeech}, CVSS \cite{jia2022cvss}, CoVoST2 \cite{wang2020covost}, and COCO \cite{lin2014microsoft}. We use this fine-tuned DMLM as a pre-trained model, and further fine-tune it on desired tasks such as multi-modal ASR. We extend the length-normalized tri-modal loss function proposed in \cite{trinh2024discrete} to accept more modalities. The model is trained using AdamW \cite{loshchilovdecoupled} with $\beta$=(0.9, 0.999), weight decay of 1e-4, lr=1e-6, on a single NVIDIA A100 GPU with a batch size of 4 and patience of 5. 

\subsection{Evaluation Metrics}
We report the Word Error Rates (WER) for each dataset, using the Whisper English text normalizer \cite{radford2023robust} to clean and standardize the text by removing extraneous characters, normalizing spaces, and converting text to lowercase. We also report a relative WER benefit of adding extra modalities to the audio task, which is calculated by $(\textrm{WER}_{A}-\textrm{WER}_{X+A})/\textrm{WER}_{A}$, where $A$ denotes audio and $X$ stands for any additional modalities. This metric indicates better performance with a larger value.


\section{Results and Discussion}
\label{sec:res}

\subsection{Do additional modalities always help ASR accuracy?}
\label{subsec:exp1}

\begin{table*}[t]
\centering
\caption{Evaluation of WER (\%) and benefit of adding image(I), lip hypothesis(L), and different types of OCR(O) on the 3-equations 2-noise test set, at different noise levels based on SNR. +$\infty$ represents clean audio and $-\infty$ means pure noise.} 
\vspace{-2em}
\begin{center}

\resizebox{\textwidth}{!}{%
\begin{tabular}{lcccccccccc}
\noalign{\hrule height 1.2pt}
\multicolumn{1}{l|}{} & (clean) & \multicolumn{7}{c}{2-noise, SNR(dB) =} & (noise) & \multicolumn{1}{|c}{}\\
\multicolumn{1}{l|}{Task} & +$\infty$ & 20 & 10 & 5 & 0 & -5 & -10 & -20 & $-\infty$ & \multicolumn{1}{|c}{\bf Average}\\ \hline

\multicolumn{11}{c}{\cellcolor[HTML]{DEDCDC}WER $\downarrow$ (relative benefit $\uparrow$)} \\

\multicolumn{1}{l|}{Whisper-base.en} & 3.88& 4.23& 4.71& 5.32& 6.89& 17.55& 49.78& 92.01& 123.4 & \multicolumn{1}{|c}{31.4}\\
\multicolumn{1}{l|}{Whisper-small.en} & 1.46& 1.36& 1.48& 1.96& 2.95& 7.96& 26.77& 55.57& 135.2 & \multicolumn{1}{|c}{23.7}\\
\multicolumn{1}{l|}{Whisper-base.en + OCR prompt} & 7.79& 8.60& 8.84& 11.84& 18.85& 44.63& 81.58& 146.40& 293.48 & \multicolumn{1}{|c}{64.0}\\
\multicolumn{1}{l|}{Whisper-small.en + OCR prompt} & 6.72& 5.48& 4.16& 4.20& 5.70& 11.08& 34.99& 87.04& 302.95 & \multicolumn{1}{|c}{46.8}\\
\hline

\multicolumn{1}{l|}{A$\rightarrow$T (average of 2 trails)} & 0.21 & 0.31 & 0.57 & 1.10 & 3.05 & 12.73 & 26.29 & 37.68 & 200.75 & \multicolumn{1}{|c}{28.5} \\
\hline

\multicolumn{1}{l|}{I+A$\rightarrow$T} & 0.21 (0.0\%) & 0.34 (-11.5\%) & 0.59 (-4.4\%) & 1.10 (0.0\%) & 2.97 {\color[HTML]{009503}(+2.6\%)} & 13.05 (-2.5\%) & 26.78 (-1.9\%) & 38.07 (-1.0\%) & 238.1 (-18.6\%) & \multicolumn{1}{|c}{32.3 (-3.5\%)} \\
\multicolumn{1}{l|}{L+A$\rightarrow$T} & 0.27 (-28.6\%) & 0.45 (-47.5\%) & 0.76 (-34.5\%) & 1.24 (-12.7\%) & 3.12 (-2.3\%) & 12.43 {\color[HTML]{009503}(+2.4\%)} & 24.71 {\color[HTML]{009503}(+6.0\%)} & 35.14 {\color[HTML]{009503}(+6.7\%)} & 191.8 {\color[HTML]{009503}(+4.5\%)} & \multicolumn{1}{|c}{27.2 (-10.5\%)} \\
\multicolumn{1}{l|}{I+L+A$\rightarrow$T} & 0.19 {\color[HTML]{009503}(+9.5\%)} & 0.35 (-14.8\%) & 0.62 (-9.7\%) & 1.18 (-7.3\%) & 3.05 (0.0\%) & 12.69 {\color[HTML]{009503}(+0.3\%)} & 25.21 {\color[HTML]{009503}(+4.1\%)} & 35.61 {\color[HTML]{009503}(+5.5\%)} & 142.6 {\color[HTML]{009503}(+29.0\%)} & \multicolumn{1}{|c}{\textbf{22.3 {\color[HTML]{009503}(+1.3\%)}}} \\
\hline

\multicolumn{1}{l|}{O+A$\rightarrow$T} & 0.24 (-14.3\%) & 0.28 {\color[HTML]{009503}(+8.2\%)} & 0.51 {\color[HTML]{009503}(+9.7\%)} & 0.86 {\color[HTML]{009503}(+21.8\%)} & 2.4 {\color[HTML]{009503}(+21.3\%)} & 11.68 {\color[HTML]{009503}(+8.3\%)} & 24.33 {\color[HTML]{009503}(+7.5\%)} & 34.66 {\color[HTML]{009503}(+8.0\%)} & 172.4 {\color[HTML]{009503}(+14.1\%)} & \multicolumn{1}{|c}{24.9 {\color[HTML]{009503}(+11.4\%)}} \\ 
\multicolumn{1}{l|}{O+L+A$\rightarrow$T} & 0.22 (-4.8\%) & 0.29 {\color[HTML]{009503}(+4.9\%)} & 0.51 {\color[HTML]{009503}(+9.7\%)} & 1.01 {\color[HTML]{009503}(+8.2\%)} & 2.57 {\color[HTML]{009503}(+15.7\%)} & 11.72 {\color[HTML]{009503}(+7.9\%)} & 24.13 {\color[HTML]{009503}(+8.2\%)} & 34.03 {\color[HTML]{009503}(+9.7\%)} & 135.9 {\color[HTML]{009503}(+32.3\%)} & \multicolumn{1}{|c}{\textbf{21.2 {\color[HTML]{009503}(+12.5\%)}}} \\ 
\multicolumn{1}{l|}{O+I+A$\rightarrow$T} & 0.17 {\color[HTML]{009503}(+19.1\%)} & 0.27 {\color[HTML]{009503}(+11.5\%)} & 0.36 {\color[HTML]{009503}(+36.3\%)} & 0.85 {\color[HTML]{009503}(+22.7\%)} & 2.38 {\color[HTML]{009503}(+22.0\%)} & 10.87 {\color[HTML]{009503}(+14.6\%)} & 23.72 {\color[HTML]{009503}(+9.8\%)} & 34.10 {\color[HTML]{009503}(+9.5\%)} & 296.4 (-47.63\%) & \multicolumn{1}{|c}{37.04 \textbf{{\color[HTML]{009503}(+12.5\%)}}} \\ 
\hline

\multicolumn{1}{l|}{$\textrm{O}_{\textrm{oracle,3}}$+A$\rightarrow$T} & 0.04 {\color[HTML]{009503}(+81.0\%)} & 0.06 {\color[HTML]{009503}(+80.3\%)} & 0.1 {\color[HTML]{009503}(+82.3\%)} & 0.14 {\color[HTML]{009503}(+87.3\%)} & 0.31 {\color[HTML]{009503}(+89.8\%)} & 1.42 {\color[HTML]{009503}(+88.9\%)} & 2.12 {\color[HTML]{009503}(+91.9\%)} & 3.14 {\color[HTML]{009503}(+91.7\%)} & 91.95 {\color[HTML]{009503}(+54.2\%)}& \multicolumn{1}{|c}{\textbf{9.9 {\color[HTML]{009503}(+83.9\%)}}} \\ 
\multicolumn{1}{l|}{$\textrm{O}_{\textrm{oracle,10}}$+A$\rightarrow$T} & 0.07 {\color[HTML]{009503}(+66.7\%)} & 0.08 {\color[HTML]{009503}(+73.8\%)} & 0.13 {\color[HTML]{009503}(+77.0\%)} & 0.32 {\color[HTML]{009503}(+70.9\%)} & 0.79 {\color[HTML]{009503}(+74.1\%)} & 5.78 {\color[HTML]{009503}(+54.6\%)} & 12.32 {\color[HTML]{009503}(+53.1\%)} & 18.79 {\color[HTML]{009503}(+50.1\%)} & 109.4 {\color[HTML]{009503}(+45.5\%)}& \multicolumn{1}{|c}{14.8 {\color[HTML]{009503}(+63.9\%)}}\\ 

\noalign{\hrule height 1.2pt}
\end{tabular}
}
\vspace{-1.5em}

\label{tab:3eq_SNR}
\end{center}
\end{table*}

On the one hand, using multiple modalities could improve accuracy by supplying the model with complementary information. On the other hand, it could conceivably hurt performance, as the additional modalities result in longer input sequences, which might prevent the model from finding relevant information. We thus conduct a modality ablation study to investigate whether fusing more modalities with audio can improve recognition performance.  In this experiment we examine the potential benefits of incorporating multiple modalities by averaging across different noise levels.

{\bf 3-Equations}:  In Table~\ref{tab:3eq_SNR} we list the WER and relative benefit of adding each modality to the audio-only baseline (A$\rightarrow$T). To obtain greater statistical reliability, we train the audio-only model twice using the shuffled training data, and calculate the average WERs of these two trails as the baseline.  Models are fine-tuned on the 3-Equations 2-noise training set, such that each audio is augmented with random MUSAN noise at an SNR in [+$\infty$, 20, 10, 5, 0, -5, -10, -20, -$\infty$]. Then the models are evaluated on 2-noise constant SNR test sets in each of [+$\infty$, 20, 10, 5, 2.5, 0, -5, -10, -20, -$\infty$]. 

Compared to the audio-only baseline, the average benefits of adding a single modality of either image (I+A$\rightarrow$T), lip (L+A$\rightarrow$T), or OCR (O+A$\rightarrow$T) are -3.5\%, -10.5\%, and +11.4\%, respectively. Hence, only the addition of OCR brings a consistent benefit across noise levels. However, when considering 3-modality combinations, we observe more consistent benefits: adding both image and lip (I+L+A$\rightarrow$T), the model surpasses the audio-only model by +1.3\%; adding both OCR and lip (O+L+A$\rightarrow$T), the model achieves an average benefit of +12.5\%; and adding both OCR and image (O+I+A$\rightarrow$T), the model also achieves an average benefit of +12.5\%. 

To ensure that visual cues are not sufficient to complete this task, we also include experiments for OCR-only (O$\rightarrow$T), lip-only (L$\rightarrow$T), and image-only (I$\rightarrow$T).  The WERs of O$\rightarrow$T, L$\rightarrow$T, and I$\rightarrow$T are 77.3, 41.8, and 79.7, respectively. When comparing O$\rightarrow$T to  O+A$\rightarrow$T, it's easy to figure out that the model needs audio information at all noise levels except the pure noise scenes. The overall WER increases by 52.4 in the absence of audio information. Same conclusion holds for the lip-only model and the image-only model, with an overall WER increment of 47.4 and 14.6. The performance gap between the single modality models and the 2-modalities models reflects the importance of integrating more modal information for such tasks.

In addition to varying the input modalities, we explore different representations of visual modality, from implicit to explicit. We consider tokens generated by the image encoder (DALL-E) as {\em implicit} visual representation, OCR as {\em explicit}, and oracle OCR as {\em accurate explicit}. As summarized in Table~\ref{tab:3eq_SNR}, the average benefits of these models follow the order: I+A$<$O+A$<\textrm{O}_{\textrm{oracle,3}}$+A. This can be attributed to the visual representation becoming more explicit and accurate, making it easier for the model to use, which suggests that better visual representation can lead to better supplementary performance.

We also compare our model to two Whisper models that are of similar size as ours. Although our model is only trained on about 4k hours of audio (per-training and fine-tuning combined), which is less than 1\% of Whisper's training set, it shows better performance by harnessing multi-modal capabilities. This suggests leveraging more modalities can optimize performance even with a limited amount of data. Since Whisper models are able to accept prompts, we also consider comparing to the prompted Whisper (as shown in Table~\ref{tab:3eq_SNR}, Whisper + OCR prompt). However, the performance degraded when prompted with OCR texts, we suspect this is because Whisper models suffer more from hallucination problem when provided with a prompt.

\begin{table}[t]
\centering
\caption{Evaluation of WER (\%) and benefit of adding OCR(O) on the SlideAVSR dataset, at different noise levels based on SNR and FQ Ranker $K^{\mathrm{a}}$ values. $\textrm{O}_{\textrm{ALL}}$ means using all OCR words.} 
\resizebox{\columnwidth}{!}{%

\begin{tabular}{lccccc}
\noalign{\hrule height 1.2pt}
\multicolumn{1}{l|}{} & (clean) & \multicolumn{2}{c}{SNR(dB) =} &  & \multicolumn{1}{|c}{} \\
\multicolumn{1}{l|}{Task} & +$\infty$ & 10 & 0 & -10 & \multicolumn{1}{|c}{Average} \\ \hline
\multicolumn{6}{c}{\cellcolor[HTML]{DEDCDC}WER $\downarrow$ (relative benefit $\uparrow$)} \\
\multicolumn{1}{l|}{A$\rightarrow$T} & 33.8 & 42.5 & 44.8 & 70.6 & \multicolumn{1}{|c}{47.9} \\
\multicolumn{1}{l|}{$\textrm{O}_{\textrm{ALL}}$+A$\rightarrow$T} & 31.4 {\color[HTML]{009503}(+7.0\%)} & 37.2 {\color[HTML]{009503}(+12.4\%) }& 47.7 (-6.4\%) & 75.9 (-7.6\%) & \multicolumn{1}{|c}{48.0 {\color[HTML]{009503}(+1.4\%)}} \\
\multicolumn{1}{l|}{$\textrm{O}_{K=30}$+A$\rightarrow$T} & 30.5 {\color[HTML]{009503}(+9.7\%)} & 35.7 {\color[HTML]{009503}(+15.8\%)} & 46.9 (-4.7\%) & 75.5 (-6.9\%) & \multicolumn{1}{|c}{\textbf{47.2} {\color[HTML]{009503}(+3.5\%)}}\\
\multicolumn{1}{l|}{$\textrm{O}_{K=10}$+A$\rightarrow$T} & 30.6 {\color[HTML]{009503}(+9.5\%)} & 34.6 {\color[HTML]{009503}(+18.5\%)} & 46.2 (-3.1\%) & 77.9 (-10.3\%) & \multicolumn{1}{|c}{47.3 {\color[HTML]{009503}(\textbf{+3.6\%})}} \\ 
\noalign{\hrule height 1.2pt}
\multicolumn{6}{l}{$^{\mathrm{a}}$Indicating maximum word counts for OCR.}
\end{tabular}%

\label{tab:slideavsr_res}
}
\vspace{-1.5em}
\end{table}

{\bf SlideAVSR}: Based on previous 3-Equations experiments, image data with more text tends to perform better with explicit representation. Therefore, we similarly conduct OCR-based experiments on SlideAVSR. As shown in Table~\ref{tab:slideavsr_res}, adding extra OCR modality ($\textrm{O}_{\textrm{ALL}}$+A$\rightarrow$T) shows an average relative benefit of +1.4\%, which confirms that adding OCR improves overall recognition performance.

\subsection{Does each modality provide a uniform accuracy boost?}
\label{subsec:exp2}

We explored the hypothesis that the benefits of multimodality would be largest when there is a ``medium'' amount of noise because (a) if the audio is very clean, the other modalities are unnecessary and (b) if the audio is very noisy,  the model cannot find a correspondence between the auditory and visual channels. Results on both datasets are described below.

\begin{figure}[t]
    \centering
    \includegraphics[width=0.86\columnwidth]{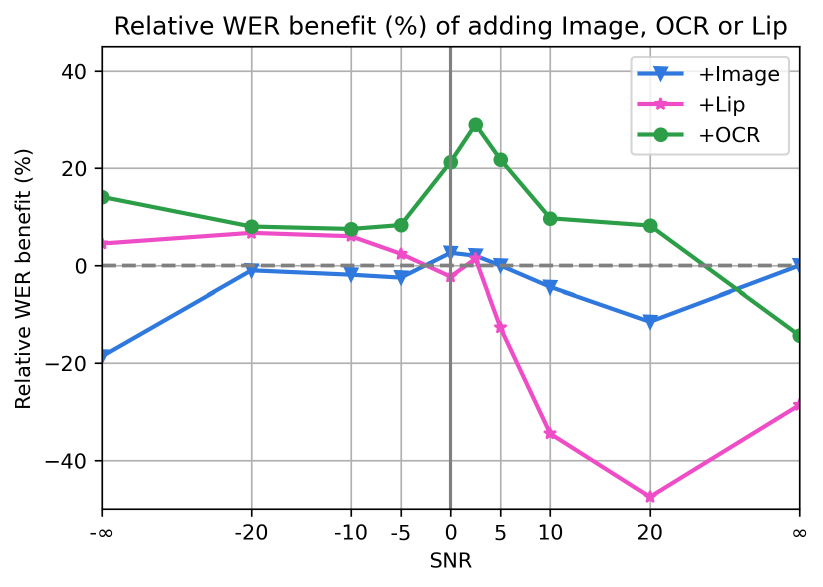}
    \vspace{-1em}
    \caption{Relative WER benefit (\%) of adding single image, OCR, or lip modalities on 3-Equations 2-noise test set. 
    }
    \vspace{-1.2em}
    \label{fig:benefit_3eq_1}
\end{figure}

{\bf 3-Equations}: In Table~\ref{tab:3eq_SNR}, when audio is clean, neither image nor lip helps improve recognition performance. However, when looking into higher noise levels, the relative benefit of adding images exhibits a very different trend compared to adding lips. As shown in Fig.~\ref{fig:benefit_3eq_1}, with increasing noise levels, the benefit of lips in enhancing accuracy becomes more amplified, showing the same trend as discussed in previous works \cite{ma2023auto}. Conversely, the benefit of images follows a trend of first increasing and then decreasing, peaking in the middle when SNR=0dB. This discrepancy could be attributed to image modality not being inherently synchronized with speech, unlike lip movements. Therefore, when above a ``sweet spot'', the audio is too noisy for the image to establish a reliable correspondence, thus the relative benefit begins to decline. The benefit of OCR shows a similar trend as image modality, but the sweet spot is likely different from the image due to visual information quality (in this case, SNR=2.5dB). 

\begin{figure}[t]
    \centering
    \includegraphics[width=0.86\columnwidth]{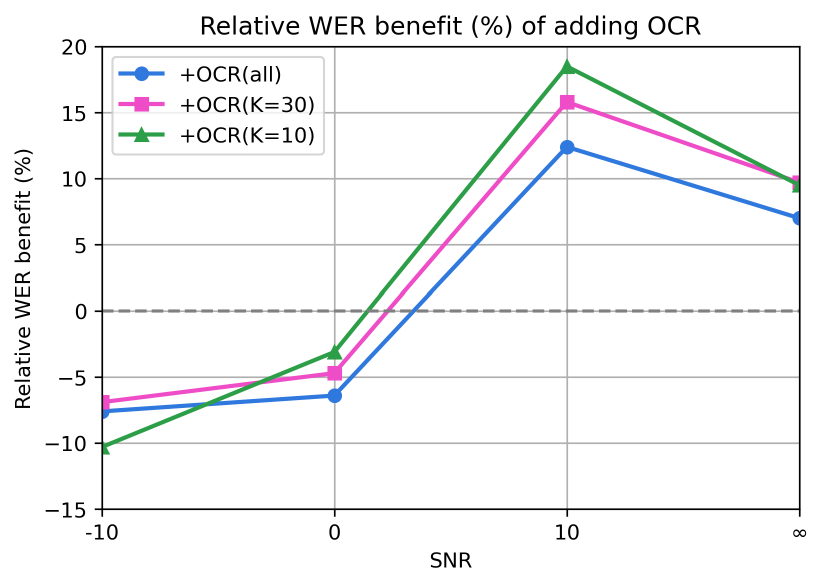}
    \vspace{-1em}
    \caption{Relative WER benefit (\%) of adding OCR with different K values for FQ Ranker on SlideAVSR test set.}
    \vspace{-1.2em}
    \label{fig:benefit_slideavsr}
\end{figure}

{\bf SlideAVSR}: The results in Table~\ref{tab:slideavsr_res} and Fig.~\ref{fig:benefit_slideavsr} show a similar phenomenon: Including the  OCR modality ($\textrm{O}_{\textrm{ALL}}$+A$\rightarrow$T) outperforms the audio-only baseline at low noise levels ($\infty$, 10dB), but performs worse at higher noise levels. The relative benefit of adding OCR initially increases, then decreases, achieving the greatest benefit of +12.4\% at SNR=10dB.

\subsection{How does irrelevant visual info.~affect ASR performance?}
\label{subsec:exp3}

Since 3-Equations dataset has 3 written equations but only 2 spoken equations in each example, the OCR inputs inherently contain 1/3 irrelevant information. Therefore, we are also interested in how the proportion of irrelevant visual inputs affects performance. On the 3-Equations dataset, we thus add 7 extra irrelevant oracle OCR sentences in addition to the 3 relevant sentences, making a dataset with 4/5 irrelevant inputs. The result is included in Table~\ref{tab:3eq_SNR} (O$_{\textrm{oracle,10}}$+A$\rightarrow$T). We observe that the overall benefit is much worse than using accurate oracle OCR. This trend confirms that adding more irrelevant visual information will hinder the model from finding the correct information, especially in noisy environments. 

\begin{figure}[t]
    \centering
    \includegraphics[width=\columnwidth]{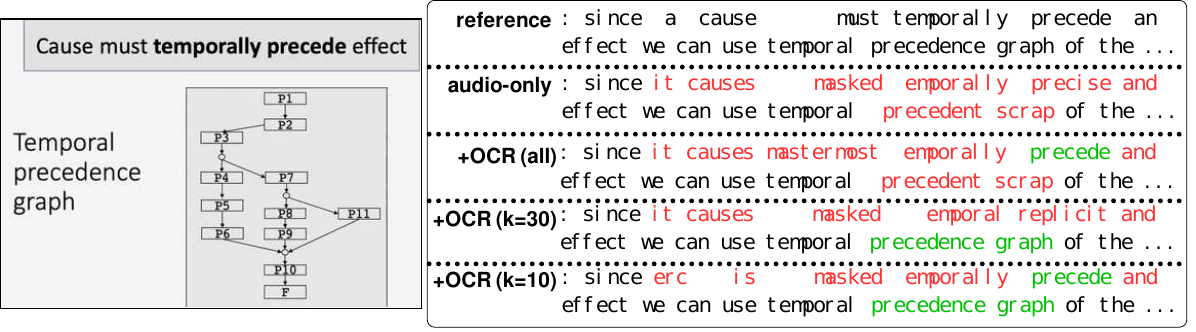}
    \vspace{-1.5em}
    \caption{An example on SlideAVSR of how filtering relevant information helps the accuracy. The corresponding slide is on the left. Red words are wrong predictions, and green words are those that were wrongly predicted by the audio-only model but corrected by the OCR plus audio model.}
    \vspace{-1em}
    \label{fig:example_slideavsr}
\end{figure}

In SlideAVSR, one slide sample can contain hundreds of OCR words, but only a small proportion is relevant to the speech. Following the work in \cite{wang2024slideavsr}, we used FQ Ranker that calculates word ranks based on the frequency of word occurrences in English Wikipedia, and filters the OCR words based on word frequency. We use 10 and 30 as the maximum word count ($K$) for prompts. This preprocessing step helps us filter the most relevant or long-tail words in OCR words, and conceivably helps the recognition performance. An example of how this filtering process helps the performance is shown in Fig.~\ref{fig:example_slideavsr}. 
As shown in Table~\ref{tab:slideavsr_res} and Fig.~\ref{fig:benefit_slideavsr}, although the performance of adding OCR is worse than that of audio-only at some noise levels, the relative benefit of OCR generally increases as we filter more stringently (i.e., smaller $K$).

\section{Conclusions}
\label{sec:conclusions}
We investigated how multiple modalities impact the accuracy of speech recognition performed by decoder-only discrete speech models. Our experiments suggest that fusing multiple modalities generally enhances recognition performance, but with caveats:
Image information exhibits a different trend from lip movements. Typically, as the noise level increases, the accuracy benefit of the lip information grows larger, whereas images provide the greatest benefit at moderate noise levels. Also, we observe a steady performance improvement when relevant visual information is filtered in preprocessing.
To our knowledge, this paper is the first to show the benefit of combining audio, images, and lip movements in one model. 

However, a limitation of our work is that using a synthesized dataset may introduce bias due to inherent synthesis errors. In future research, other modality backbone models, such as VQ-Wav2Vec and ViViT, and more real-world datasets should be explored to assess the generalizability of our findings. 


{\bf Acknowledgment}:
This research was supported by the NSF National AI Institute for Student-AI Teaming (iSAT) under grant DRL \#2019805, and also from an NSF CAREER grant \#2046505. The opinions expressed are those of the authors and do not represent views of the NSF.


\small

\bibliographystyle{IEEEbib}
\bibliography{refs}

\end{document}